\documentclass[lettersize,journal,twoside]{IEEEtran}

\IEEEoverridecommandlockouts                              

\usepackage{cite}
\usepackage{amsmath,amssymb,amsfonts}
\usepackage{array}
\usepackage[caption=false,font=normalsize,labelfont=sf,textfont=sf]{subfig}
\usepackage{stfloats}
\usepackage{url}
\usepackage{verbatim}
\def\BibTeX{{\rm B\kern-.05em{\sc i\kern-.025em b}\kern-.08em
    T\kern-.1667em\lower.7ex\hbox{E}\kern-.125emX}}
\usepackage{balance}
\usepackage{algorithmic}
\usepackage{graphicx}
\usepackage{textcomp}
\usepackage{array}
\usepackage{booktabs}
\usepackage{rotating}
\usepackage{xcolor}
\usepackage{multirow}

\begin{document}
\title{CertainNet: Sampling-free Uncertainty Estimation for Object Detection}
\author{Stefano Gasperini$^{*}$ \quad Jan Haug$^{*}$ \quad Mohammad-Ali Nikouei Mahani \quad Alvaro Marcos-Ramiro\\Nassir Navab \quad Benjamin Busam \quad Federico Tombari
\thanks{Manuscript received: September, 9, 2021; Accepted: November, 7, 2021.}
\thanks{This paper was recommended for publication by Editor Cesar Cadena upon evaluation of the Associate Editor and Reviewers' comments.}
\thanks{
This work was partly sponsored by the German Federal Ministry for Economic Affairs and Energy (grant number: 19A19005B) through the VDA KI-Absicherung project.}
\thanks{$^{*}$ The authors contributed equally.}
\thanks{Stefano Gasperini and Jan Haug are with the Faculty of Computer Science, Technische Universit\"{a}t M\"{u}nchen (TUM), 85748 Garching bei M\"{u}nchen, Germany, and also with the BMW Group, 80788 M\"{u}nchen, Germany (corresponding author {\tt\footnotesize  stefano.gasperini@tum.de, jan.haug@tum.de}).}
\thanks{Mohammad-Ali Nikouei Mahani was with the BMW Group when this work was conducted, 80788 M\"{u}nchen, Germany ({\tt\footnotesize  man.mahani@gmail.com}).}
\thanks{Alvaro Marcos-Ramiro is with the BMW Group, 80788 M\"{u}nchen, Germany ({\tt\footnotesize  alvaro.marcos-ramiro@bmw.de}).}
\thanks{Nassir Navab is with the Faculty of Computer Science, Technische Universit\"{a}t M\"{u}nchen (TUM), 85748 Garching bei M\"{u}nchen, Germany, and also with the Department of Computer Science, Johns Hopkins University, Baltimore, 21218 MD USA ({\tt\footnotesize nassir.navab@tum.de}).}
\thanks{Benjamin Busam is with the Faculty of Computer Science, Technische Universit\"{a}t M\"{u}nchen (TUM), 85748 Garching bei M\"{u}nchen, Germany ({\tt\footnotesize b.busam@tum.de}).}
\thanks{Federico Tombari is with the Faculty of Computer Science, Technische Universit\"{a}t M\"{u}nchen (TUM), 85748 Garching bei M\"{u}nchen, Germany, and also with Google, 8002 Z\"{u}rich, Switzerland ({\tt\footnotesize tombari@in.tum.de}).}
\thanks{This is a preprint of the article accepted at Robotics and Automation Letters (RA-L). \copyright~2021 IEEE.}
}

\markboth{IEEE Robotics and Automation Letters. Preprint Version. Accepted November, 2021}
{Gasperini \MakeLowercase{\textit{et al.}}: CertainNet: Sampling-free Uncertainty Estimation for Object Detection}  

\maketitle

\begin{abstract}
Estimating the uncertainty of a neural network plays a fundamental role in safety-critical settings. In perception for autonomous driving, measuring the uncertainty means providing additional calibrated information to downstream tasks, such as path planning, that can use it towards safe navigation. In this work, we propose a novel sampling-free uncertainty estimation method for object detection. We call it CertainNet, and it is the first to provide separate uncertainties for each output signal: objectness, class, location and size. To achieve this, we propose an uncertainty-aware heatmap, and exploit the neighboring bounding boxes provided by the detector at inference time. We evaluate the detection performance and the quality of the different uncertainty estimates separately, also with challenging out-of-domain samples: BDD100K and nuImages with models trained on KITTI. Additionally, we propose a new metric to evaluate location and size uncertainties. When transferring to unseen datasets, CertainNet generalizes substantially better than previous methods and an ensemble, while being real-time and providing high quality and comprehensive uncertainty estimates.
\end{abstract}


\section{INTRODUCTION}

\IEEEPARstart{W}{hile} neural networks have been widely employed in research and mobile applications, their usage is still limited in safety-critical real-world settings, such as autonomous driving~\cite{gawlikowski2021survey}. The ability of a model to estimate the uncertainty of its predictions is a key enabler for its safe and reliable use in these contexts and unknown scenarios~\cite{feng2018towards}.
This makes uncertainty estimation a fundamental companion to object detection~\cite{choi2019gaussian}, semantic segmentation~\cite{postels2019sampling}, visual odometry~\cite{yang2020d3vo}, and other relevant visual tasks, especially to handle out-of-domain data.

\begin{figure}[t]
\centering
  \includegraphics[width=0.485\textwidth]{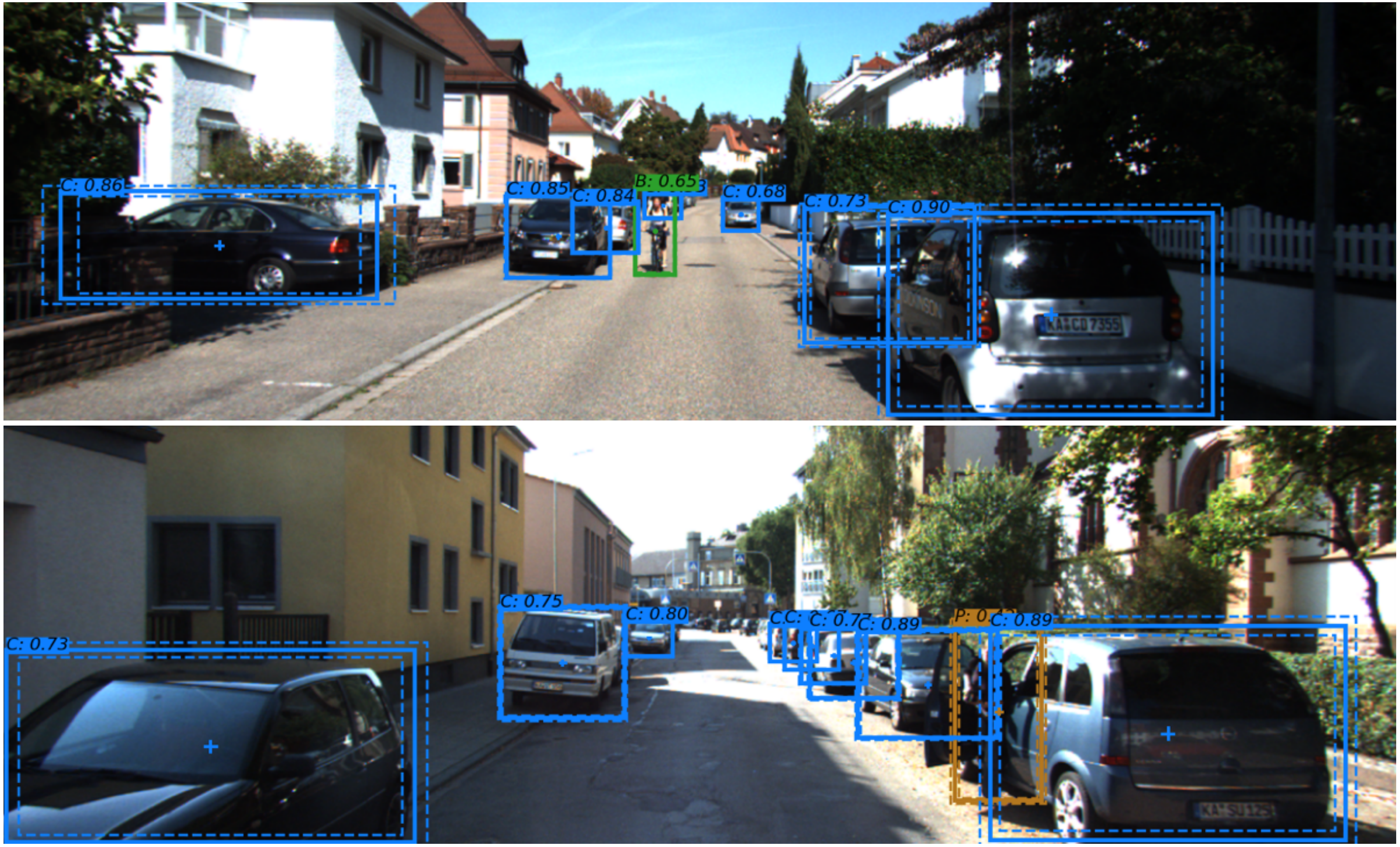}
   \caption{
   Example predictions of the proposed CertainNet from the KITTI validation set~\cite{Geiger2012CVPR}.
   Uncertainties for objectness, location and size are shown as the complement of the value at the top left of each box, the plus around the center, and dashed boxes respectively. The object class is color-coded.
   }
   \label{fig:teaser}
\end{figure}

As one of the main perception tasks, object detection comprises a plurality of outputs (e.g., object location, size, and class), requiring both regression and classification, and rendering it particularly challenging for uncertainty estimation.
Existing real-time approaches have focused on providing an uncertainty estimate only on a subset of these detection outputs (e.g., location and size~\cite{choi2019gaussian,lee2020localization}). 
In traffic, there can be situations where an object's location and size are relatively certain, while its existence in the scene (i.e., objectness) is uncertain, e.g., a person in a dark area or entering a car with an unusual pose, as shown at the bottom of Figure~\ref{fig:teaser}. In these cases, it is particularly important to assess the objectness uncertainty, to avoid ignoring the detection solely based on a low confidence score.
A method confidence is a score which does not correspond to the actual probability of being correct, while uncertainty reduces this gap~\cite{gawlikowski2021survey}.

However, prior works cover this only partially, as they leave behind the uncertainty over important parameters, such as the objectness, posing a potential safety issue.
By extracting a dedicated uncertainty measure for each of the output signals, a system would provide deeper and better insights on the scene, which could be exploited by downstream tasks (e.g., path planning).
Furthermore, incorporating uncertainty can reduce the accuracy of the models~\cite{cortinhal2020salsanext}, or significantly increase the latency~\cite{gal2016dropout, miller2019evaluating}, hence leading to a challenging trade-off between accuracy, safety, and runtime.

In this work, we address these issues with a novel method that provides separate and dedicated uncertainties for all detection outputs, while preserving speed and accuracy. We name our approach CertainNet, and our contributions can be summarized as follows:
\begin{itemize}
    \item We introduce a novel sampling-free method to estimate the model uncertainty of an object detector.
    \item We propose the first real-time approach to estimate the uncertainty of all signals of the detection output (i.e., objectness, location, size and class).
    \item We present new simple metrics to assess the quality of size and location uncertainties for object detection.
\end{itemize}
Additionally, we provide extensive evaluations and comparisons on three challenging public datasets, including out-of-domain data, reaching competitive accuracy, while improving significantly on uncertainty-aware metrics.


\section{RELATED WORK}

\subsection{Uncertainty Estimation for Neural Networks}
A learning-based system can express two kinds of uncertainty: the epistemic uncertainty caused by the model, and the aleatoric uncertainty due to the input data~\cite{gawlikowski2021survey}.
The former can arise from errors in the training procedure (e.g., ignoring data imbalance), a sub-optimal architecture or the knowledge gap with out-of-domain data, while the latter originates from the data itself, e.g., due to sensor noise, coarse approximations, ambiguous samples, or also inaccurate ground truth labels~\cite{gawlikowski2021survey}.

There are several types of uncertainty estimation approaches. In this Section we provide a brief overview following the categorization of Gawlikowski et al.~\cite{gawlikowski2021survey}. \textbf{Single deterministic methods} predict with a deterministic model and provide an uncertainty estimate directly through that model~\cite{sensoy2018evidential}, or via external computation~\cite{van2020uncertainty, raghu2019direct}. There are various ways for external methods to estimate the uncertainty, e.g., by computing the distance to the training data. DUQ, which is the work of Van Amersfoort et al.~\cite{van2020uncertainty}, belongs to this subcategory. The authors used Radial Basis Functions (RBF) to learn a transformation of features into a hyperspace. At inference time, they quantify the uncertainty as the distance in the hyperspace between the predicted features and a set of learned class centroids.

Another group of works is that of \textbf{Bayesian methods}, which infer the probability distribution of the model parameters~\cite{gawlikowski2021survey}. In this category, Monte-Carlo Dropout (MC Dropout)~\cite{gal2016dropout} approximates this distribution by considering the set of models originating from keeping dropout active at test time. Thanks to the randomness of dropout, each forward pass produces different outputs, from which the uncertainty can be estimated according to their disagreement.

Then, \textbf{ensemble methods} also combine multiple outputs for both predictions and uncertainty estimates~\cite{lakshminarayanan2017simple}. In this case the outputs are inferred by a set of different deterministic models, trained independently. Subcategories of ensembles can be found depending on how these different models are obtained~\cite{gawlikowski2021survey}.

Lastly, \textbf{test-time augmentation methods} use a single deterministic model and augment its input during inference, collecting a variety of predictions, from which a final output and uncertainty estimate are computed~\cite{wang2019aleatoric}.

A different categorization distinguishes sampling-based and sampling-free approaches, depending on how the uncertainty is estimated. The former group requires inferring multiple times on the same input with different models (for Bayesian and ensemble methods), or different inputs with the same model (with test-time augmentations), and then aggregating the results. This significantly increases the runtime~\cite{gal2016dropout, miller2019evaluating}, which makes it impractical for most real-time scenarios, such as autonomous driving. Instead, the latter category does not require sampling, thanks to built-in quantification techniques, such as those typically used in single deterministic methods~\cite{van2020uncertainty}.

Our work builds on top of the findings of Van Amersfoort et al.~\cite{van2020uncertainty}. In particular, we considerably extend and adapt their DUQ from image classification to a dense regression task for object detection, estimating the model objectness uncertainty at a given location, while significantly improving training stability and convergence. Additionally, we estimate the uncertainty on the variety of regressed outputs of object detection (e.g., object size), by aggregating the predictions provided by the detector.

\subsection{Uncertainty Estimation in Object Detection}
While many approaches quantify the uncertainty on basic tasks, posing the foundations of this domain~\cite{gal2016dropout, van2020uncertainty}, fewer estimated it for safety-critical settings, such as autonomous driving~\cite{postels2019sampling, harakeh2020bayesod}.
As we focus on estimating the uncertainty for object detection, in this Section we provide a brief summary of existing works in this area.

Choi et al. proposed Gaussian YOLOv3 (GYOLO)~\cite{choi2019gaussian}, extending the popular YOLOv3 detector~\cite{redmon2018yolov3} to estimate the object location and size uncertainties. They achieved this by predicting Gaussian parameters (i.e., mean and variance) for the position and dimensions of the boxes.
Miller et al.~\cite{miller2019evaluating} used MC Dropout~\cite{gal2016dropout} and thoroughly evaluated different clustering techniques, extracting variance and entropy of spatial and class predictions respectively.
Harakeh et al.~\cite{harakeh2020bayesod} also used MC Dropout~\cite{gal2016dropout} and replaced the usual non-maximum suppression (NMS) with clustering and Bayesian inference, obtaining an uncertainty quantification for the object size and class.
Instead, Lee et al.~\cite{lee2020localization} introduced with their Gaussian-FCOS a dedicated head on top of an anchor-less detector to estimate the localization uncertainty for each of the four box boundaries independently.

Likewise, our proposed CertainNet estimates the uncertainty for object detection, but in a different way. Unlike the works of Miller et al.~\cite{miller2019evaluating} and Harakeh et al.~\cite{harakeh2020bayesod}, ours is sampling-free and real-time capable. Compared to GYOLO~\cite{choi2019gaussian} and Gaussian-FCOS~\cite{lee2020localization}, ours does not regress the uncertainty as an explicit additional output, thereby reducing the potential impact of biases in the training data.
Moreover, we are the first to quantify the uncertainty for each individual aspect of the detection output (i.e., objectness, location, size and class), increasing the explainability.

\begin{figure}[t]
\centering
  \includegraphics[width=0.47\textwidth]{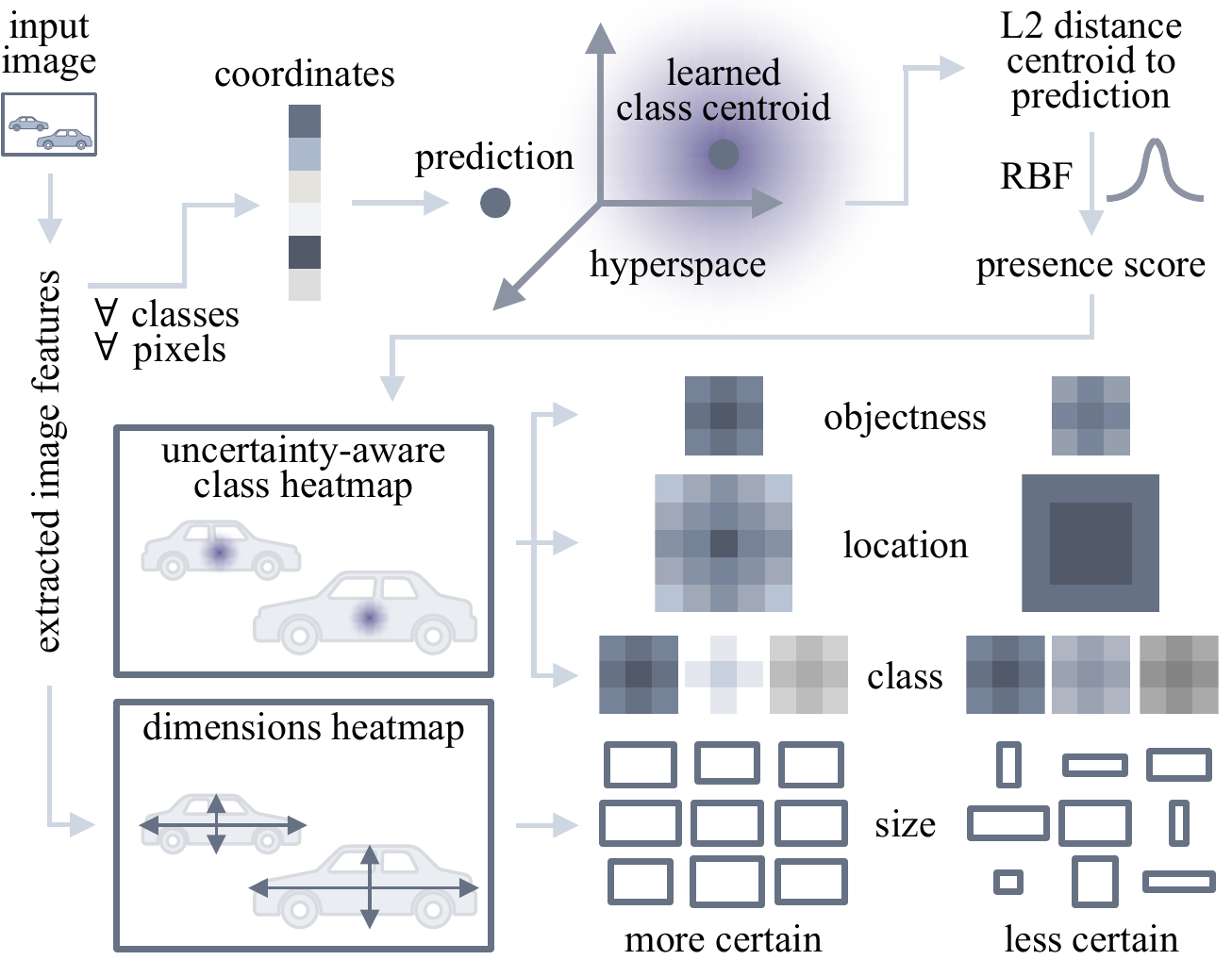}
   \caption{The proposed CertainNet. The top shows how each uncertainty-aware objectness score is computed. At the bottom right, examples of more and less certain predictions for each output signal are shown.
   }
   \label{fig:small_framework}
\end{figure}

\section{METHOD}
In this work, we estimate the uncertainty for each aspect of object detection, improving the model explainability for safety-critical applications.
Towards this end, we build on top of an anchor-less detection framework (Section~\ref{sec:detection_framework}), which we render uncertainty-aware (Section~\ref{sec:uncertainty_awareness}). We then compute the uncertainty for objectness, location, dimensions and class  (Section~\ref{sec:uncertainties}). Figure~\ref{fig:small_framework} shows an overview of the proposed method.
This fragmentation of the uncertainty along the individual detection components introduces valuable information to improve the safety of downstream tasks.

\subsection{Base Detection Framework}\label{sec:detection_framework}
CenterNet~\cite{zhou2019objects} is a fast and accurate anchor-less detector. It predicts objects centers via a class heatmap, and regresses the bounding box size in a separate dimensions heatmap. CenterNet gives as intermediate output a bounding box for every pixel, hence providing a distribution of outputs to estimate our uncertainties. For these reasons, we use it as base detector.
While we focus on 2D bounding boxes, 3D boxes can be predicted following~\cite{zhou2019objects}, and the additional uncertainties (e.g., for object distance and length) can be computed analogously to the ones presented in this work.

\subsection{Uncertainty-awareness}
\label{sec:uncertainty_awareness}
We make the class heatmap of the detector uncertainty-aware, by exploiting DUQ~\cite{van2020uncertainty}, extending it from recognizing out-of-domain samples in image classification tasks, for which it was proposed, to a dense regression task within our object detector.
During training, we learn a set of class representatives (i.e., centroids), which are then compared with each prediction at inference time. By doing so, similarly to DUQ~\cite{van2020uncertainty}, we assess the deviation from learned object prototypes, which is linked to the model uncertainty.
We compute this for all regressed heatmap values, which represent the objectness scores. In particular, it is the comparison with the learned centroids, which makes every value of this heatmap uncertainty-aware.

\begin{figure}[t]
\centering
  \includegraphics[width=0.47\textwidth]{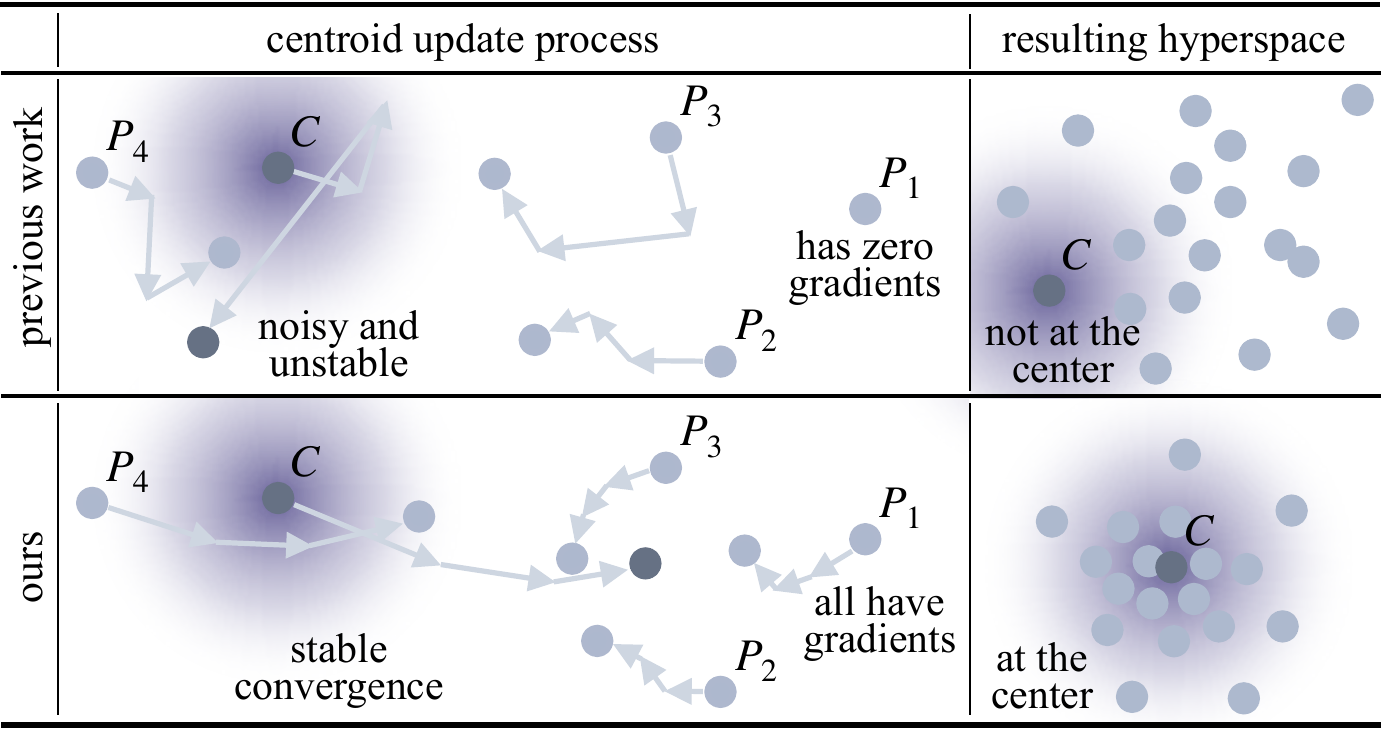}
   \caption{Centroid update scheme, with the learned hyperspace. The bottom shows the impact of our modifications, compared to the unstable procedure of DUQ~\cite{van2020uncertainty} at the top when applied on an unbalanced detection task.
   }
   \label{fig:centroid_update}
\end{figure}

Proposed for image classification, DUQ~\cite{van2020uncertainty} learns to map each image to $N_{cls}$ high dimensional spaces, one for each of $N_{cls}$ classes. At training time, class centroids are updated as moving averages of positive samples.
As Berger et al.~\cite{berger2021confidence} also pointed out, DUQ suffers from instability issues and deployment difficulties.
Therefore, after performing a thorough analysis of DUQ and its feature space, we incorporate several modifications to better suit the object detection task, as well as to improve the overall convergence and stability of the model, as visualized in Figure~\ref{fig:centroid_update}.

\textbf{Adaptation of DUQ to dense regression}:
Compared to the original implementation~\cite{van2020uncertainty}, due to the high dimensionality of the centroid space (e.g., 512 in~\cite{van2020uncertainty}), and also the two additional dimensions required for our class heatmap output (i.e., width and height), speed and memory efficiency issues arise. We circumvent these by exploiting the shared operation across the pixels when
transforming to the hyperspace. Towards this end, we use $1 \times 1$ convolutions, instead of individual multiplications as in~\cite{van2020uncertainty}.

\textbf{Balanced centroid update}:
The centroids need to be a good representation of the data, as at inference time they determine both predictions and uncertainties, making the centroid update a critical step.
In particular, the centroids are updated as trailing averages, of which the current estimates are computed to get closer to the current sets of predictions. In a regular setup~\cite{van2020uncertainty}, scaling by the amount of samples of a given class $c$ is reasonable, as each minibatch $t$ is representative of the whole dataset. However, with unbalanced classes and a varying amount of objects, that would lead to inconsistent update magnitudes, hindering convergence.
We circumvent this and account for the minibatch variability by computing the centroid $\textbf{e}_{c,t}$ as:
\begin{equation}\label{eq:centroid_updates}
    \mathbf{e}_{c,t} = \gamma \cdot \mathbf{e}_{c,t-1} + \frac{(1-\gamma)}{\sum_i \mathbf{y}_{c,t,i}^\lambda} \sum_i \left( \mathbf{y}_{c,t,i} ^\lambda \circ \mathbf{W}_c f_\theta (\mathbf{x}_{t,i}) \right)
\end{equation}
This allows to set a weight on the center pixel higher than its surroundings, depending on $\lambda$. The right term is an average weighted and scaled by the ground truth class heatmap values $\mathbf{y}_{c,t}$, having a Gaussian around each object center.
$\mathbf{W}_c f_\theta (\mathbf{x}_{t,i})$ are the predicted hyperspace coordinates, with $f_\theta$ being the feature extractor, and $\mathbf{W}_c$ the hyperspace transformation for class $c$. $\gamma$ is a hyperparameter, defined as the centroid momentum, as in ~\cite{van2020uncertainty}.

\textbf{Hyperspace regularization}:
To properly represent the training data, the centroids should lie among positive samples distributed along a Gaussian hypersphere. However, by visualizing the hyperspace via PCA and t-SNE, we noticed that this is not the case, with the centroid often falling around the boundary of an irregularly shaped training distribution (top of Figure~\ref{fig:centroid_update}).
We circumvent this and aim at the ideal case, by regularizing this hyperspace with:
\begin{equation}
    \mathcal{L}_{reg} = \frac{1}{\sum_{c,i} \mathbf{y}_{c,i}^\lambda} \sum _{c,i}  \left( \mathbf{y}_{c,i} ^\lambda \circ ||\mathbf{W}_c f_\theta (\mathbf{x}_{i}) - \mathbf{e}_c||_2 ^2 \right)
\end{equation}
which acts on the euclidean distance between the centroid and prediction for each pixel, and is weighted by the ground truth class heatmap values $\mathbf{y}_{c,i}$.
Furthermore, we do not use the gradient penalty $\mathcal{L}_{grad}$ of DUQ~\cite{van2020uncertainty}, as we find $\mathcal{L}_{reg}$ to deliver more stable and consistent results.

\textbf{Centroid momentum scheduling}:
The centroids are moved using steps with a magnitude proportional to their distance to the predictions, and to the centroid momentum $\gamma$.
A large $\gamma$ slows down training, while improving its stability, but reduces the robustness against the initialization. Instead, a low $\gamma$ has the inverse effect. While a relatively low $\gamma$ worked well for balanced classes and large batch sizes~\cite{van2020uncertainty}, it did not in our task.
We apply a scheduling of $\gamma$ similar to that of learning rates, to improve the stability in later training stages, and preserve robustness and speed at earlier ones. This effect is represented in Figure~\ref{fig:centroid_update}.
Specifically, we reduce $\gamma$ by a factor of 10 at fixed epochs.

\textbf{Outliers protection}:
Additionally, we prevent outliers from impacting the centroids. In particular, we define an outlier as a prediction being further from its centroid than the triple of the length scale $\sigma$.
This further improves the training stability and convergence, as instead of having the predictions following a highly moving target (i.e., centroid), they go towards a more stable target.

\textbf{Length scale annealing}:
At early stages of training, the centroids position and the hyperspace transformation are not yet properly tuned, due to random initialization. This, together with the high dimensionality of the space, causes instability, as even the correct inputs can be mapped far from their centroid.
We alleviate this problem and improve stability with an approach similar to that of simulated annealing~\cite{kirkpatrick1983optimization}: we increase the length scale $\sigma$ at the initial stages, to then slowly reduce it at each step towards the original value. This increases the gradients at larger distances, nearly zero otherwise, thereby allowing to improve convergence and predictions, as depicted in Figure~\ref{fig:centroid_update}.

\begin{table*}[h!]
\caption{Detection performance comparison with related works on the KITTI~\cite{Geiger2012CVPR} validation set across classes and difficulties. 
}
\begin{center}
\begin{tabular}{l|c|ccc|ccc|ccc|cc|r}
\toprule
 & & \multicolumn{3}{c|}{\textit{Car}} & \multicolumn{3}{c|}{\textit{Pedestrian}} & \multicolumn{3}{c|}{\textit{Cyclist}} & \multicolumn{2}{c|}{all}\\
Method & mAP & Easy & Mod. & Hard & Easy & Mod. & Hard & Easy & Mod. & Hard & AUPR-In & AUPR-Out & fps \\
\midrule
GaussianYOLO~\cite{choi2019gaussian} & 56.74 & 89.91 & 84.36 & 75.65 & 58.43 & 50.95 & 43.69 & 45.97 & 31.17 & 30.53 & 65.39 & 96.42 & \underline{18.06} \\
CenterNet~\cite{zhou2019objects} & 71.49 & 92.30 & 89.15 & \textbf{82.17} & \underline{76.53} & \underline{67.53} & \textbf{59.37} & 73.48 & 52.63 & 50.25 & 68.33 & \underline{99.82} & \textbf{20.21} \\
5-Ensemble~\cite{miller2019evaluating} & \textbf{73.70} & \textbf{97.84} & \textbf{89.75} & 80.87 & \textbf{77.08} & \textbf{67.66} & \underline{59.17} & \textbf{79.45} & \underline{57.30} & \underline{54.17} & \textbf{72.13} & 97.32 & 5.10\\
CertainNet~[ours] & \underline{73.00} & \underline{93.81} & \underline{89.36} & \underline{82.11} & 76.33 & 66.13 & 58.54 & \underline{78.02} & \textbf{57.49} & \textbf{55.20} & \underline{69.85} & \textbf{99.85} & 16.46 \\
\bottomrule
\end{tabular}%
\label{tab:KITTI_mAP}
\end{center}
\end{table*}

\begin{table*}[h!]
\caption{
Uncertainty quality comparison with related works on the validation sets of KITTI~\cite{Geiger2012CVPR}, BDD100K~\cite{yu2020bdd100k}, and nuImages (\textnormal{nuIm.})~\cite{caesar2020nuscenes}. All models were trained on KITTI, and transferred ($\rightarrow$) to the other datasets. All results are for \textit{Car} (moderate for KITTI). On \textnormal{CenterNet$^\ddag$} we applied our location and dimensions uncertainty estimations as post-processing.}
\begin{center}
\begin{tabular}{c|l|c|cccrr|rcc|rcc}
\toprule
& & & \multicolumn{5}{c|}{objectness} & \multicolumn{3}{c|}{location} & \multicolumn{3}{c}{dimensions}  \\
 & Method & AP & AUPR-In & AUPR-Out & AUROC & ECE $\downarrow$ & UE $\downarrow$ & CE $\downarrow$ & UBQ & BR & CE $\downarrow$ & UBQ & BR\\
\midrule

\parbox[t]{2mm}{\multirow{4}{*}{\rotatebox[origin=c]{90}{KITTI}}}
& GaussianYOLO~\cite{choi2019gaussian} & 84.36 & \underline{74.93} & 91.24 & 86.23 & 23.37 & 20.02 & 11.58 & 60.01 & 60.56 & 6.18 & 75.34 & 77.47\\
& CenterNet$^\ddag$~\cite{zhou2019objects} & 89.15 & 72.76 & \underline{99.56} & \underline{96.65} & \underline{10.49} & \underline{6.00} & 7.17 & 68.27 & 69.80 & \textbf{4.23} & \underline{85.84} & \underline{89.58}\\
& 5-Ensemble~\cite{miller2019evaluating} & \textbf{89.75} & 73.79 & 93.61 & 87.13 & 16.93 & 17.72 & \underline{4.59} & \textbf{86.71} & \textbf{90.43} & 5.01 & 84.28 & 87.54\\
& CertainNet~[ours] & \underline{89.36} & \textbf{78.16} & \textbf{99.81} & \textbf{98.00} & \textbf{4.60} & \textbf{4.98} & \textbf{4.54} & \underline{75.92} & \underline{77.81} & \underline{4.47} & \textbf{86.76} & \textbf{91.05} \\
\midrule

\parbox[t]{2mm}{\multirow{4}{*}{\rotatebox[origin=c]{90}{$\rightarrow$ BDD}}}
& GaussianYOLO~\cite{choi2019gaussian} & 31.03 & \underline{88.73} & 79.72 & 85.84 & 22.80 & 20.68 & \underline{9.48} & \underline{57.32} & \underline{58.49} & \textbf{7.67} & 57.94 & 59.19\\
& CenterNet$^\ddag$~\cite{zhou2019objects} & \underline{34.51} & 81.35 & \textbf{98.06} & \underline{91.74} & \underline{7.78} & \textbf{14.03} & 25.72 & 35.90 & 36.76 & 10.53 & \underline{73.43} & \underline{79.55}\\
& 5-Ensemble~\cite{miller2019evaluating} & 26.75 & \textbf{96.42} & 81.84 & \textbf{91.99} & 25.64 & 15.26 & \textbf{3.31} & \textbf{84.83} & \textbf{93.54} & \underline{9.40} & \textbf{84.81} & \textbf{93.50}\\
& CertainNet~[ours] & \textbf{40.93} & 78.77 & \underline{97.82} & 91.17 & \textbf{4.89} & \underline{14.85} & 14.79 & 44.71 & 46.03 & 9.68 & 72.91 & 78.89\\
\midrule

\parbox[t]{2mm}{\multirow{4}{*}{\rotatebox[origin=c]{90}{$\rightarrow$ nuIm.}}}
& GaussianYOLO~\cite{choi2019gaussian} & 30.38 & \underline{86.24} & 87.95 & 87.78 & 17.05 & 17.74 & 10.42 & \underline{56.20} & \underline{57.62} & 8.74 & 58.38 & 60.14\\
& CenterNet$^\ddag$~\cite{zhou2019objects} & \underline{43.93} & 75.94 & \underline{98.93} & 91.58 & \textbf{5.51} & 15.70 & 27.07 & 35.69 & 36.21 & 8.39 & 74.79 & 79.21 \\
& 5-Ensemble~\cite{miller2019evaluating} & 35.31 & \textbf{93.93} & 90.90 & \underline{93.04} & 17.74 & \underline{13.98} & \textbf{3.37} & \textbf{89.48} & \textbf{98.51} & \underline{8.36} & \textbf{89.31} & \textbf{98.21}\\
& CertainNet~[ours] & \textbf{53.14} & 79.97 & \textbf{99.28} & \textbf{94.24} & \underline{7.80} & \textbf{11.90} & \underline{16.21} & 43.57 & 44.46 & \textbf{7.73} & \underline{75.36} & \underline{80.25}\\

\bottomrule
\end{tabular}
\label{tab:multi_uncertainties}
\end{center}
\end{table*}

\subsection{Uncertainties Estimation}
\label{sec:uncertainties}

\textbf{Objectness}:
\label{sec:objectness}
Thanks to the class heatmap uncertainty-awareness (Section~\ref{sec:uncertainty_awareness}), the objectness uncertainty $\mathcal{U}_{obj}$ can be directly inferred as the complement of the score:
\begin{equation}\label{eq:u_obj}
    \mathcal{U}_{obj} = 1 - \hat{p} \quad \text{with:} ~ \hat{p} = \max _c p\left(\mathbf{W}_c f_\theta (\mathbf{x})\  |\   \mathcal{N}(\mathbf{e}_c, \sigma )\right)
\end{equation}
with $\mathcal{N}(\mathbf{e}_c, \sigma )$ being a Gaussian distribution centered at the centroid $\mathbf{e}_c$ with variance $\sigma$, which is the length scale.

\textbf{Location}:
\label{sec:location}
An object location is computed according to the class heatmap. In our CertainNet, this output is uncertainty-aware (Section~\ref{sec:uncertainty_awareness}). We make use of this by extracting the location uncertainty considering the distribution of scores around an object center. As can be seen in Figure~\ref{fig:small_framework}, a location is more certain if the center peak is more defined and sharper, while it is less certain with a smoother and flatter peak. The uncertainty $\mathcal{U}_{x}$ for the location in $x$ direction depends on the horizontal variance and is normalized by the predicted width $\hat{w}$. We compute $\mathcal{U}_{x}$ as follows:
\begin{equation}
    \mathcal{U}_{x} = \frac{\sqrt{\textrm{var}_x}}{\hat{w}}
    \quad \text{with:} ~ 
    \textrm{var}_x = \frac{\sum_i o_x^2(i) ~\hat{p}_i \cos^\eta{\alpha_i}}{\sum_i \hat{p}_i \cos^\eta{\alpha_i}}
\end{equation}
where $i$ iterates over the pixels around the center, $o_x(i)$ is the x-offset from pixel $i$ to the center, $\alpha_i$ is the angle between the x-axis and the pixel, and $\hat{p}_i$ is the predicted objectness (Equation~\ref{eq:u_obj}). $\eta$ weights the influence of the scores that are not axis-aligned.
The vertical location uncertainty $\mathcal{U}_{y}$ is computed analogously, replacing $\hat{w}$ with the predicted height $\hat{h}$ and using $\textrm{var}_y$, obtained through the y-offset $o_y(i)$ and $\sin{\alpha_i}$.

\textbf{Dimensions}:
\label{sec:dimensions}
As described in Section~\ref{sec:detection_framework}, the model provides a bounding box for each pixel in the output heatmaps. We exploit this by computing the uncertainty on the object size $\mathcal{U}_{w,h}$, by comparing the boxes around an object center. Intuitively, the more these boxes are similar to one another, the more certain the model is on the object dimensions, and vice versa, as shown in Figure~\ref{fig:small_framework}.
Towards this end, we compute $\mathcal{U}_{w}$ as an RMSE over the surrounding predicted dimensions for each object center, and weight this by the pixel objectness score:
\begin{equation}
    \mathcal{U}_{w} = \frac{1}{\hat{w}}\sqrt{\frac{1}{\sum_i \hat{p}_i} \sum_i \hat{p}_i \cdot (\hat{w} - \hat{w}_i)^2}
\end{equation}
where $\hat{w}$ is the predicted object width and $\hat{w}_i$ is the width at the heatmap pixel $i$ around the center. The height uncertainty $\mathcal{U}_{h}$ is computed analogously using the predicted heights $\hat{h}$.

\textbf{Class}:
\label{sec:class}
The object class is predicted via the class heatmap, which is uncertainty-aware in our model (Section~\ref{sec:uncertainty_awareness}). As in Figure~\ref{fig:small_framework}, a class prediction is more certain, if there is a peak only at such class with the other class probabilities being low, while it is less certain, the higher those other probabilities are.
We formalize this, by comparing the objectness scores at each predicted center across the classes:
\begin{equation}
    \mathcal{U}_{cls}(i) = \mathcal{U}_{cls}(i-1)+(1-\mathcal{U}_{cls}(i-1))\cdot \left(\frac{\hat{p}_{cls}(i)}{\hat{p}_{cls}(1)}\right)^{i}
\end{equation}
with $\hat{p}_{cls}(i)$ being the $i^{th}$-highest class probability, so $i=1$ for the predicted one. We compute $\mathcal{U}_{cls}$ recursively, considering $\mathcal{U}_{cls}(0)=0$ and reaching the final class uncertainty $\mathcal{U}_{cls}$ as $\mathcal{U}_{cls}(N_{cls})$.

\section{EXPERIMENTS AND RESULTS}

\subsection{Experimental setup}

\textbf{Datasets}:
We conducted our experiments on three public autonomous driving datasets, namely KITTI~\cite{Geiger2012CVPR}, the recent nuImages extension of nuScenes~\cite{caesar2020nuscenes}, and the dashcam-based BDD100K~\cite{yu2020bdd100k}. \textbf{KITTI} is a popular benchmark for object detection, and it has been recorded in Germany. We followed the standard 3DOP split~\cite{zhou2019objects}, which comprises 3712 training and 3769 validation images.
Moreover, we used the standard \textit{car}, \textit{pedestrian} and \textit{cyclist} classes. \textbf{NuImages} is a highly diverse large-scale dataset, collected in Boston and Singapore. It
includes rain, snow and night conditions.
\textbf{BDD100K} is another diverse large-scale dataset. It was crowd-sourced from a variety of dashboard cameras and different vehicles, around cities of the USA. It also includes various challenging weather and lighting conditions, such as snow, fog and night.
In particular, we used nuImages and BDD100K to assess the generalization capability to out-of-domain data, which is critical for autonomous driving and uncertainty estimation. Towards this end, we evaluated KITTI models (without any fine-tuning) on the validation set of nuImages, which includes 3249 images, sized 1600x900, and the validation set of BDD100K, which contains 10K samples, sized 1280x720. Due to the sub-optimal class overlap between KITTI and the two transfer datasets, for the evaluation we focused on the \textit{car} class.

\begin{figure}[t]
\centering
  \includegraphics[width=0.47\textwidth]{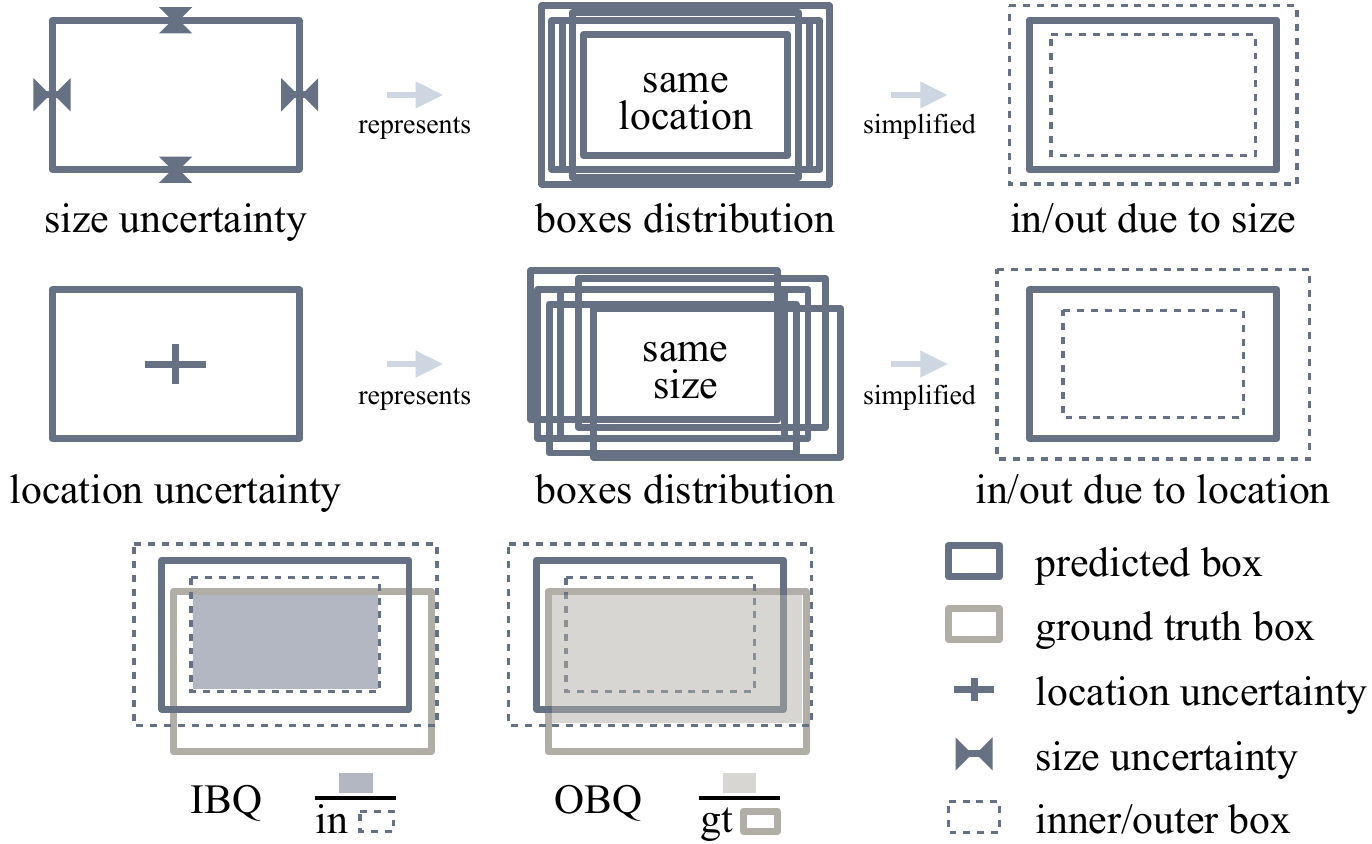}
   \caption{ 
   The top shows the effect of size and location uncertainties on the box uncertainty boundaries, allowing to evaluate them in a similar fashion. The bottom shows the IBQ and OBQ terms of our novel UBQ metric.
   }
   \label{fig:location_size_box_uncertainties}
\end{figure}

\textbf{Evaluation metrics}:
We evaluated the object detection performance on the standard AP, with the standard IoU threshold of 0.7 for KITTI \textit{cars} and 0.5 for the other classes and datasets.
We split the evaluation according to the specific uncertainty estimates, which we thoroughly evaluated.
For the objectness uncertainty we computed various metrics, such as the area under the precision-recall (AUPR) curve, as AUPR-In and AUPR-Out~\cite{miller2019evaluating}, which should be considered jointly. The former assesses the ability to accept all correct detections and minimise wrong acceptances. Conversely, the latter looks at the rejection of negative samples and the rate of wrong rejections. Additionally, we computed the area under the receiver operating characteristic curve (AUROC), comparing true and false positives rates, as used by Miller et al.~\cite{miller2019evaluating}. Following~\cite{guo2017calibration}, we calculated the expected calibration error (ECE) as the average difference between confidence and detection accuracy, weighted by the occurrence. Moreover, we evaluated the minimum uncertainty error (UE) as the ability to accept correct detections and reject incorrect ones based on the uncertainty estimate~\cite{miller2019evaluating}.
In the top of Figure~\ref{fig:location_size_box_uncertainties}, we show how location and size uncertainties have a similar effect on the box uncertainty boundaries, despite leading to different box distributions. This allows to evaluate them in the same way. We adapted the expected calibration error (ECE)~\cite{guo2017calibration} for our detection case, as the calibration error (CE): average deviation between predicted uncertainty and measured error for matched detections. We independently computed the CE on \textit{x} and \textit{y} for the location, and on \textit{width} and \textit{height} for the size, reporting the average of each pair.
All metrics in this work are percentage-based.

\textbf{Uncertainty boundary quality}:
We introduce a new metric to evaluate the uncertainties on the object location and size. As shown at the bottom of Figure~\ref{fig:location_size_box_uncertainties}, the goal is assessing how much of an object is in the uncertainty boundaries.
We name this metric uncertainty boundary quality (UBQ), and compute it as the average across matched boxes of:
\begin{equation}
    UBQ_i = \frac{1}{2}\left( IBQ_i  +  OBQ_i \right) BR_i
\end{equation}
with BR being the boundary ratio between inner and outer:
\begin{equation}
    BR_i = \frac{1}{2} \left( \frac{\text{width}(in_i)}{\text{width}(out_i)} + \frac{\text{height}(in_i)}{\text{height}(out_i)} \right)
\end{equation}
UBQ is composed of an inner (IBQ) and an outer (OBQ) term, which are averaged and multiplied by the ratio (BR), to penalize extremely broad and trivial confidence intervals. Figure~\ref{fig:location_size_box_uncertainties} shows how $IBQ_i$ and $OBQ_i$ are computed:
\begin{equation}
    IBQ_i = \frac{A(gt_i \cap in_i)}{A(in_i)}\quad\quad OBQ_i = \frac{A(gt_i \cap out_i)}{A(gt_i)}
\end{equation}
where $A$ is the area of a bounding box $i$, $gt$ is the ground truth box, $in$ and $out$ are the inner and outer predictions.
As for the CE, our UBQ metric can be used for both location and size. We report UBQ as a summary, and BR as the ratio.

\textbf{Network architecture}:
All our models were based on a DLA architecture~\cite{yu2018deep}. We selected it as compared to others it offers a good speed-accuracy trade-off~\cite{zhou2019objects}. Thus, we followed the DLA-34 configuration of CenterNet, with deformable convolutions~\cite{zhou2019objects}. We then made the class heatmap head uncertainty-aware, with the hyperspace transformation and the RBF kernel, as described in Section~\ref{sec:uncertainty_awareness}. We used a single model for all three classes of KITTI.

\begin{figure*}[t]
\centering
  \includegraphics[width=0.98\textwidth]{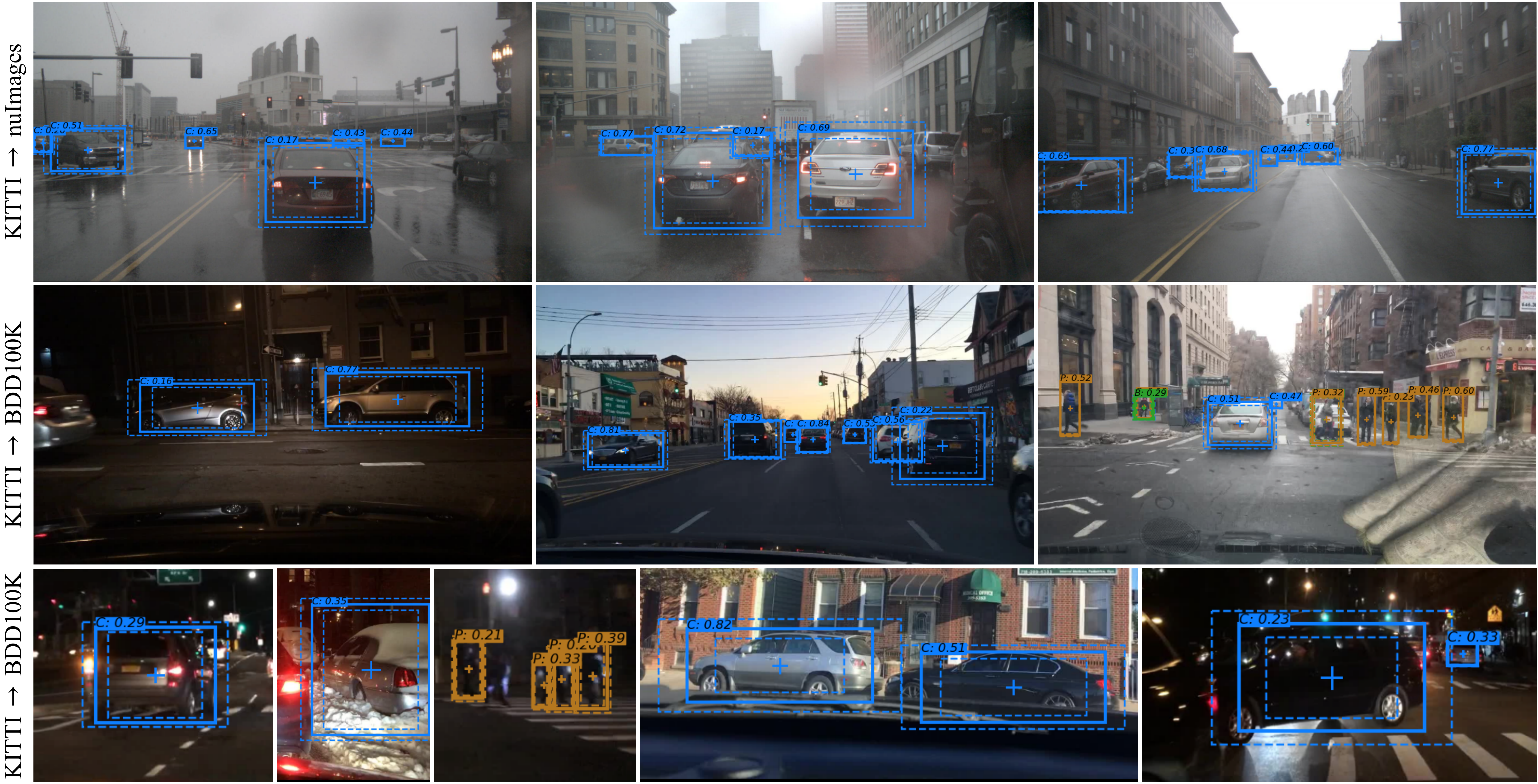}
   \caption{
   Example predictions of the proposed CertainNet on out-of-domain data. The model was trained only on KITTI~\cite{Geiger2012CVPR} and applied on nuImages~\cite{caesar2020nuscenes} and BDD100K~\cite{yu2020bdd100k} without any fine-tuning. The bottom row shows crops. The boxes follow the format introduced in Figure~\ref{fig:teaser}, with KITTI $\rightarrow$ KITTI.
   }
   \label{fig:qualitative}
\end{figure*}

\textbf{Implementation details}:
We trained our models with an Adam optimizer for $70$ epochs, with $10$ additional epochs for the objectness head. We used an initial learning rate of $1.25e-4$, reduced by a factor of $10$ after epochs $45$ and $60$. We chose a batch size of 16 and a resolution of $1280\text{x}384$ for KITTI, resizing BDD and nuImages samples to $896\text{x}512$. We selected $512$ as centroid dimensionality, and initialized their momentum $\gamma$ to $0.9$, increasing it to $(0.99, 0.999, 0.9999)$ at epochs $(5, 20, 60)$ respectively. The length scale $\sigma$ was initialized at $2.5e-1$, with a decay rate of $0.999$ per step for the sigma annealing, until reaching $5e-2$.
We weighted the hyperspace regularization loss $\mathcal{L}_{reg}$ by $1e-2$, and set $\lambda$ to $20$ for balancing the score. To compensate the added loss on the objectness, the weight on the dimensions loss was increased to $0.2$. We set $\eta=4$, and kept the remaining hyperparameters the same as proposed for CenterNet~\cite{zhou2019objects}.
We trained our models using PyTorch on a single NVIDIA Tesla V100 32GB GPU and evaluated the runtime (fps) on a single NVIDIA Quadro RTX 4000 8GB GPU.

\begin{table}[t]
\caption{Ablation study on the KITTI~\cite{Geiger2012CVPR} validation set according to \textit{Car} moderate detection and uncertainty quality.
}
\begin{center}
\begin{tabular}{ll|ccc}
\toprule
ID & Description & AP & AUPR-In & AUPR-Out \\
\midrule
A0 & adapted DUQ~\cite{van2020uncertainty} & 76.41 & \textbf{81.64} & 99.62 \\
A1 & A0 - $\mathcal{L}_{grad}$ + $\mathcal{L}_{reg}$ & 87.33 & 76.26 & 99.15 \\
A2 & A1 + balanced update & 87.98 & 74.12 & 99.71 \\
A3 & A2 + outlier protection & 88.14 & 77.34 & 99.56 \\
A4 & A3 + momentum schedul. & 88.76 & 75.27 & 99.72 \\
A5 & A4 + length scale anneal. & 88.85 & 81.08 & 99.80 \\
A6 & A5 + freeze last 10 epochs & \textbf{89.36} & 78.16 & \textbf{99.81}\\
\bottomrule
\end{tabular}%
\label{tab:KITTI_ablations}
\end{center}
\end{table}

\textbf{Details for prior works}:
For a fair comparison, we re-trained all methods on the same standard dataset split~\cite{zhou2019objects}, using the official implementations and hyperparameters, starting from weights pretrained on ImageNet~\cite{russakovsky2015imagenet}, until convergence.
For GYOLO~\cite{choi2019gaussian} we used the best settings with input size 704 on Darknet.
When evaluating location and size uncertainties, we extended CenterNet~\cite{zhou2019objects} with our location and dimensions uncertainty estimations, on top of its outputs.
For the ensemble, we trained 5 DLA-34-based CenterNets (randomly initialized, with data shuffle and augmentations), and combined their predictions with the merging strategy of Miller et al.~\cite{miller2019evaluating}, via DBSCAN.
For CenterNet, GYOLO and the ensemble, the objectness uncertainty is evaluated using the standard confidence scores provided.

\subsection{Quantitative Results}
\label{sec:quantitative}

\textbf{Detection}:
Table~\ref{tab:KITTI_mAP} shows the detection performance of our approach compared to related works on KITTI~\cite{Geiger2012CVPR}. Our uncertainty-aware CertainNet outperformed the CenterNet~\cite{zhou2019objects} baseline on the overall mAP, as well as most classes and difficulties, while sharing the same architecture. Our additional safety-related outputs reduced the runtime by 18\% to 16.46 fps, thereby remaining real-time.
Moreover, the sampling-based 5-Ensemble~\cite{miller2019evaluating} models achieved the overall highest mAP, and the best score in most classes and difficulties, with ours securing a close second place. However, the superior performance of the ensemble came at the cost of a significantly higher runtime.
GYOLO~\cite{choi2019gaussian} was slightly faster than ours, but produced the worst detections among tested methods. As shown in Table~\ref{tab:KITTI_mAP}, its AP scores are significantly lower than those reported by Choi et al. in~\cite{choi2019gaussian}, due to the different dataset splits used for KITTI. In particular, Choi et al. split training and validation randomly across single frames. We used instead a standard split~\cite{zhou2019objects}, reducing the similarity between training and validation sets.

\textbf{Uncertainty estimation}:
In Table~\ref{tab:multi_uncertainties} we report a variety of uncertainty metrics for different datasets, for the main class \textit{Car}.
On KITTI, our CertainNet achieved higher scores and lower errors than the other methods across most metrics. Remarkably, our approach was able to achieve the best objectness results across the board, often with a substantial margin (e.g., on the calibration ECE), thanks to its effective uncertainty-aware heatmap.
Interestingly, in terms of uncertainties, the ensemble could not achieve satisfactory results, except for the location. This can be attributed to the disagreement of the 5 models, and to the difficulty of merging contrasting predictions in the case of object detection.
Table~\ref{tab:multi_uncertainties} shows also the suboptimal performance of GYOLO~\cite{choi2019gaussian}, which had a tendency to overestimate the uncertainty boundaries, especially for the location (BR around 60\%), despite its training time modifications for location and size, and having been trained on the same dataset.
Furthermore, the flexibility of our uncertainty quantification techniques allowed to produce estimates also for the location and size of the uncertainty-unaware CenterNet~\cite{zhou2019objects}. However, this configuration could not match the uncertainty quality of our CertainNet, especially for the location, computed on the class heatmap. This shows how our training time modifications contributed to the uncertainty-awareness of the whole model.

\textbf{Out-of-domain data}:
As all models were trained on KITTI, the results for BDD100K and nuImages in Table~\ref{tab:multi_uncertainties} show the ability of each method to generalize to unseen scenarios in out-of-domain data. As also pointed out in~\cite{gasperini2021r4dyn, guizilini20203d}, multiple differences define a substantial domain gap between these two datasets and KITTI: weather and lighting conditions, image resolution and aspect ratio, camera mounting position and type (e.g., dashcam often with reflections on BDD), different continents and street layouts, amount of dynamic scenes, as well as different vehicles on sale in those regions. All these render the transfer task particularly difficult.
Most notably, our CertainNet reached the highest AP on both datasets, by a significant margin on both BDD and nuImages. This shows the robustness of our approach, as well as the strong generalization provided by our uncertainty estimations, compared to CenterNet~\cite{zhou2019objects}.
Interestingly, the ensemble, which performed well on KITTI (i.e., the training dataset), underperformed on both out-of-domain datasets. We attribute this to the rule-based clustering step required to merge the predictions~\cite{miller2019evaluating}, which was tuned on KITTI. This reiterates that the ensemble's aggregation step is critical on object detection, and is prone to deliver suboptimal results on out-of-domain data, which is fundamental in autonomous driving.
GYOLO~\cite{choi2019gaussian} followed a similar trend, but suffered the transfer to out-of-distribution less than the ensemble, thanks to its learning-based techniques that generalized better.
Nevertheless, the uncertainty estimations of the ensemble seemed high quality for the most part, especially on the object location and size. However, this is due to the rather low number of detections that all related works provided, as discussed below.
Overall, our CertainNet achieved the highest AP, while maintaining high quality uncertainty estimates, which correctly increased on out-of-domain data.

\textbf{Amount of detections}:
The uncertainty evaluation metrics used in this work are highly influenced by the number of predictions or that of matched detections. Therefore, high scores and low error rates can be achieved by a model that outputs a low amount of high confidence boxes. We found this to be the case for the methods in Table~\ref{tab:multi_uncertainties}: in nuImages~\cite{caesar2020nuscenes} our method detected 4905 \textit{cars}, CenterNet 3714, 5-Ensemble 3229, and GYOLO 2248; similarly in BDD~\cite{yu2020bdd100k}, our CertainNet 26802, CenterNet 20210, ensemble 16440, and GYOLO 15395.
These values also show the limitations of the ensemble, confirming that contrasting predictions within its models, especially in highly uncertain scenarios, such as out-of-domain data (e.g., BDD100K and nuImages in Table~\ref{tab:multi_uncertainties}), can happen and thereby significantly degrade the overall detection and uncertainty performance of the system.
In general, each of the uncertainty metrics evaluates only a specific aspect and should be considered together with others, and combined with the detection performance (i.e., AP and mAP).

\textbf{Ablation study}:
The importance of our modifications over DUQ~\cite{van2020uncertainty} is testified by the results shown in Table~\ref{tab:KITTI_ablations}, where A6 represents our full approach.
In particular, the baseline A0, which is the adaptation of DUQ~\cite{van2020uncertainty} from image classification to object detection, could not converge properly, due to the severe limitations described in Section~\ref{sec:uncertainty_awareness}, worsened by the high imbalance of KITTI~\cite{Geiger2012CVPR}. The biggest improvement was brought by A1 replacing the gradient penalty $\mathcal{L}_{grad}$ of DUQ with our hyperspace regularization $\mathcal{L}_{reg}$, which also improved the consistency of training convergence. Moreover, the AP increased monotonously with the introduction of each of our modifications (Section~\ref{sec:uncertainty_awareness}), while maintaining good uncertainty estimates throughout, thereby showing their significant impact over DUQ (i.e., A0).


\subsection{Qualitative Results}
\label{sec:qualitative}
In Figures~\ref{fig:teaser} and~\ref{fig:qualitative} we show predictions of our method on challenging scenes of KITTI and out-of-domain samples respectively. The images confirm the large domain gap between KITTI and the two transfer datasets, as described in Section~\ref{sec:quantitative}. The night and rainy settings in Figure~\ref{fig:qualitative} from nuImages~\cite{caesar2020nuscenes} and BDD100K~\cite{yu2020bdd100k} are completely different from the sunny and overcast scenarios of KITTI~\cite{Geiger2012CVPR} on which the model was trained, making these very difficult samples. Nevertheless, the proposed CertainNet correctly detected most objects. Compared to the KITTI predictions in Figure~\ref{fig:teaser}, the objectness certainties (values at the top left of each box) are lower for the unusual and difficult to see out-of-domain vehicles in Figure~\ref{fig:qualitative}. This shows that the model recognized them as out-of-distribution samples. Interestingly, the uncertainty estimates for location and size improved the overlap between sub-optimal predictions and the object, e.g., bottom left of Figure~\ref{fig:qualitative}, as tested by our UBQ metric.

\section{CONCLUSION}
In this work we introduced CertainNet, a novel method to estimate the uncertainty of every aspect of 2D object detection. Extensive evaluations showed the benefit of quantifying the uncertainty, especially when transferring to out-of-distribution data for safety-critical applications. Our method substantially improved the generalization ability over previous works, with more accurate predictions and high quality uncertainty estimates, while running in real-time. Therefore, the proposed CertainNet constitutes a valuable contribution towards robust object detection for autonomous driving.

\bibliographystyle{IEEEtran}
\bibliography{refs}

\end{document}